\documentclass{svmult_jk}

\usepackage{mathptmx}       
\usepackage{helvet}         
\usepackage{courier}        
\usepackage{type1cm}        
\usepackage{graphicx}        
\usepackage{multicol}        
\usepackage[bottom]{footmisc}

\usepackage{bm} 

\begin{document}

\title*{Stiffness Analysis of Parallel Manipulators with Preloaded Passive Joints}
\author{A. Pashkevich, A. Klimchik and D. Chablat}
\institute{Institut de Recherche en Communications et en Cybern\'etique de Nantes, France  \\
\'Ecole des Mines de Nantes, France  \\
\email{anatol.pashkevich@emn.fr, alexandr.klimchik@emn.fr, damien.chablat@irccyn.ec-nantes.fr}}
%
%
%
\maketitle

\abstract{The paper presents a methodology for the enhanced stiffness analysis of parallel manipulators with internal preloading in passive joints. It also takes into account influence of the external loading and allows computing both the non-linear ``load-deflection'' relation and the stiffness matrices for any given location of the end-platform or actuating drives. Using this methodology, it is proposed the kinetostatic control algorithm that allows to improve accuracy of the classical kinematic control and to compensate position errors caused by elastic deformations in links/joints due to the external/internal loading. The results are illustrated by an example that deals with a parallel manipulator of the Orthoglide family where the internal preloading allows to eliminate the undesired buckling phenomena and to improve the stiffness in the neighborhood of its kinematic singularities.}


\section{Introduction}

Parallel manipulators have become very popular in many industrial applications due to their inherent advantages of providing better accuracy, lower mass/inertia properties, and higher structural rigidity compared to their serial counterparts \cite{mono:Merlet}. These features are induced by the specific kinematic structure, which eliminates the cantilever-type loading and allows to minimize deflections caused by external torques and forces. One recent development in this area, which is targeted at high-precision manipulation, is a replacing the standard passive joints by preloaded ones, which contain internal passive springs eliminating the backlash or ensure some degree of static balancing \cite{journal:Arsenault,journal:Griffis}. This modification obviously improves the manipulator performances but requires some revision of existing stiffness analysis techniques that are in the focus of this paper.   

In most of previous works, the manipulator stiffness analysis was based on the linear modeling assumptions which ignore influence of the external or internal forces \cite{journal:Ceccarelli,journal:Company,journal:Chen,journal:Alici,journal:Ciblak}. Consequently, relevant techniques are targeted at linearization of the ``force-deflection'' relation in the neighborhood of the non-loaded equilibrium, which is perfectly described by the stiffness matrix \cite{journal:Kovecses,journal:Quennouelle}. However, in the case of non-negligible internal and/or external loading, the manipulator may demonstrate essentially non-linear behaviour, which is not exposed in the unloaded case \cite{mono:Timoshenko}. In particular, the loading may potentially lead to multiple equilibriums, to bifurcations of the equilibriums or to static instability of certain manipulator configurations \cite{journal:Su,journal:Carricato}.

This paper presents an extension of our previous results \cite{journal:Pashkevich:1} devoted to the stiffness analysis of parallel manipulators by generalizing them for case of internal preloading \cite{journal:Crane} in the passive joints. It implements the virtual joint method (VJM) of Salisbary \cite{journal:Salisbury} and Gosselin \cite{journal:Gosselin:1} that describes the compliance of the manipulator elements by a set of localized multi-dimensional springs separated by rigid links and perfect joints. The proposed technique allows computing the loaded equilibrium, finding the full-scale ``load-deflection'' relation and evaluating the corresponding stiffness matrices for any given location of the end-platform or actuating drives \cite{journal:Pashkevich:2}. It is also developed a kinetostatic control algorithm that allows to improve accuracy of the classical kinematic control and to compensate position errors caused by elastic deformations in links/joints due to the external/internal loading.

The remainder of this paper is organized as follows. Section 2 defines the research problem and basic assumptions. Section 3 deals with computing of the loaded static equilibrium and corresponding ``load-deflection'' relation. Section 4 focuses on its linearization and evaluation of the stiffness matrix. Section 5 presents the kinetostatic control algorithm. Section 6 contains an illustrative example. And finally, Section 7 summarizes the main results and contributions.
\section{Manipulator model}
Let us consider a general parallel manipulator that is composed of \textit{n} serial kinematic chains connecting a fixed base and a moving platform Figure 1. It is assumed, that the chain architecture ensures kinematic control of the manipulator but may introduce some redundant constraints that improve the rigidity. Following the VJM-concept \cite{journal:Gosselin:1}, let us presents the manipulator chains as sequences of pseudo-rigid links separated by rotational or translational joints of one of the following types: (i) perfect passive joints ; (ii) preloaded passive joints that include auxiliary flexible elements; (iii) virtual flexible joints that describe compliance of the actuators and manipulator links; (iv) actuating joints. Using this notation the geometrical model of the chain may be written as 

\begin{equation}\label{Eq:1}
\mathbf{t}=\mathbf{g}(\mathbf{\rho },\mathbf{q},\mathbf{\vartheta} ,\mathbf{\theta }),
\end{equation}
where the vector $\mathbf{t}=(\mathbf{p},~\mathbf{\varphi})^{T}$ includes the Cartesian position $\mathbf{p}={(x,\,\,y,\,\,z)}^{T}$ and orientation $\mathbf{\varphi}={(\varphi_{x},\,\,\varphi_{y},\,\,\varphi_{z})}^{T}$ of the end-platform, $\mathbf{\rho}$ is the vector of actuated coordinates (they are constant for static analysis), the vector $\mathbf{q}$ contains coordinates of all perfect passive joints, the vector $\mathbf{\vartheta}$ includes coordinates of the preloaded passive joints, and the vector $\mathbf{\theta}$  collects coordinates of all virtual springs describing elasticity of the links and joints. 

\begin{figure}[t]
\center
\includegraphics[width=11.5cm]{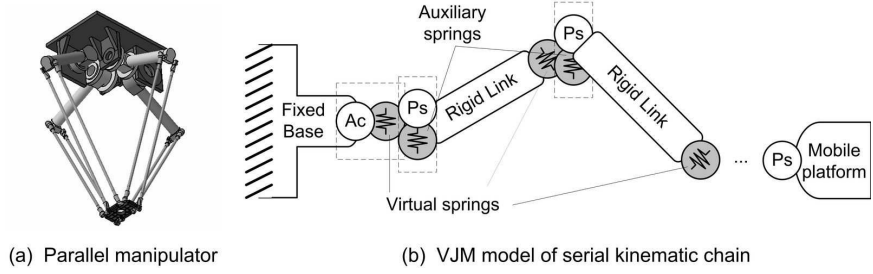}
\caption{Typical parallel manipulator and VJM-model of its kinematic chain.}
(Ac - actuator; Ps - Passive joint)
\\
\label{Figure:1}
\end{figure}

\begin{figure}[t]
\center
\includegraphics[width=11.5cm]{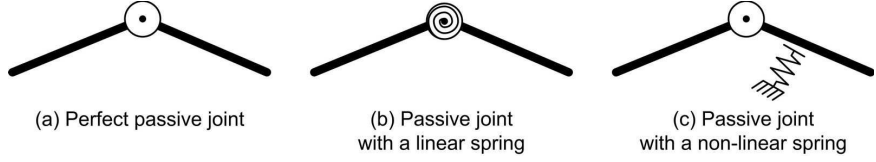}
\caption{Examples of auxiliary springs in preloaded passive joints.}
\label{Figure:2}
\end{figure}

The above mentioned elements of the kinematic chain differ in their static characteristics. In particular, the joints (i) and (iii) are described by the standard expressions \cite{journal:Pashkevich:1}

\begin{equation}\label{Eq:2}
\mathbf{\tau}_{q}=\mathbf{0}   ~~~and~~~    \mathbf{\tau}_{\theta}=\mathbf{K}_{\theta}\cdot \mathbf{\theta}
\end{equation}
where $\mathbf{\tau}_{q}$ and $\mathbf{\tau}_{\theta}$ are the generalized force/torque reactions corresponding to the aggregated vectors of the passive joint coordinates $\mathbf{q}$ and virtual joint coordinates $\mathbf{\theta }$;  $\mathbf{K}_{\theta}$ is the generalized stiffness matrix of all virtual springs. However, the preloaded passive joints (ii) may include both linear and non-linear auxiliary springs, some examples of which are shown in Figure~\ref{Figure:2}. In this paper, we will describe statics of the preloaded joints by a general expression

\begin{equation}\label{Eq:3}
\mathbf{\tau}_{\vartheta}=\mathbf{K}_{\vartheta}\cdot \mathbf{h}(\vartheta -\vartheta_{0})
\end{equation}
where $\mathbf{\tau}_{\vartheta}$ is the generalized force/torque reactions corresponding to the aggregated vectors of the preloaded joint coordinates $\mathbf{\vartheta}$; $\mathbf{\vartheta}_{0}$ defines the preloading value;  $\mathbf{K}_{\vartheta}$ is the generalized stiffness matrix of preloaded joints, and the vector function $\mathbf{h}(...)$ is assumed to be piecewise-linear, such that each of its scalar components  $h_{i}(...)$  can be expresses either as the difference $(\vartheta_{i}-\vartheta_{i0})$, or its positive or negative part  ${[\vartheta_{i}-\vartheta_{i0}]}^{+}$, ${[\vartheta_{i}-\vartheta_{i0}]}^{-}$ (see Figure 2 for details).

Using these assumptions, let us derive the stiffness model of the considered manipulator and sequentially consider the following sub-problems: (i) computing the loaded static equilibrium and obtaining the ``load-deflection'' relation; (ii) linearization of this relations in the neighborhood of this equilibrium and computing the stiffness matrix; (iii) developing the kinetostatic control algorithm, which allows to compensate position errors caused by the elastic deformations and preloading.

\section{Static equilibrium}

Let us obtain first the configuration of each kinematic chain $(\mathbf{q},\mathbf{\theta },\vartheta )$ and external force $\mathbf{F}$ that correspond to the static equilibrium with the end-point location $\mathbf{t}$. Obviously, it is a dual problem compared to the classical static analysis but it is more reasonable here because of strictly parallel structure of the considered manipulator (see Figure 1). The latter allows applying the same technique to all kinematic chains (with the same end-point location) and to compute the total external loading as the sum of the partial loadings.       

Taking into account the assumption on the piecewise-linear property of the function $\mathbf{h}(.)$, let us perform regrouping of the variables. In particular, for each current configuration of the chain, the coordinates of the preloaded passive joints described by the vector $\mathbf{\vartheta}$ may be separated into two parts $\mathbf{\vartheta}_{\theta}$ and $\mathbf{\vartheta}_{q}$ , where the first one corresponds to the active state of the auxiliary springs and the second part describes non-active springs (see Figure 2 for geometrical interpretation). This allows replacing the original set of the configuration variables $(\mathbf{q},\mathbf{\theta},\mathbf{\vartheta})$ by a set of two vectors $({\mathbf{\tilde{q}},~\mathbf{\tilde{\theta}}})$, where $\mathbf{\tilde{q}}$  aggregates the joint coordinates $(\mathbf{q},{\mathbf{\vartheta}_{q}})$ that currently are passive and the vector $\mathbf{\tilde{\theta }}$  collects all spring coordinates $(\mathbf{\theta},{\mathbf{\vartheta }_{\theta }})$ (both virtual and passive).

Using these notations and applying the virtual work technique, the static equilibrium equation of the kinematic chain may be written as

\begin{equation}\label{Eq:4}
\mathbf{\tilde{J}}_{\theta}^{T}\cdot \mathbf{F}=\mathbf{\tilde{K}}_{\theta} \cdot (\mathbf{\tilde{\theta}} - \mathbf{\tilde{\theta}}_{0});~~~~~~~~~~\mathbf{\tilde{J}}_{q}^{T}\cdot \mathbf{F}=\mathbf{0}
\end{equation}
where $\mathbf{F}$ is the external force applied at the end-point of the chain,  the vector $\mathbf{\tilde{\theta }}_{0}^{T}=[\mathbf{0}^{T},\ \mathbf{\rho}_{0}^{T}]$ aggregates the spring preloadings (which is obviously zero for the virtual springs), $\mathbf{\tilde{K}}_{\theta}={\rm diag}(\mathbf{K}_{\theta}, \mathbf{K}_{\vartheta})$, and $\mathbf{\tilde{J}}_{\theta}$, $\mathbf{\tilde{J}}_{q}$ are the kinematic Jacobians derived from (1) by differentiating it with respect to $\mathbf{\tilde{\theta }}$, $\mathbf{\tilde{q}}$. This system of equation (4) combined with the geometrical model (1), which must be rewritten in terms of the redefined variables  

\begin{equation}\label{Eq:5}
\mathbf{t}=\mathbf{\tilde{g}}(\mathbf{\tilde{q}},\mathbf{\tilde{\theta }}).
\end{equation}
This yields the desired joint coordinates of the static equilibrium for a separate kinematic chain with given end-point location.

Since the derived system is highly nonlinear, in general case a desired solution can be obtained only numerically. In this paper, it is proposed to use the following iterative scheme

\begin{equation}\label{Eq:6}
	\begin{array}{l}
 	\left[ {
 		\begin{array}{*{20}{c}}
   	\mathbf{F}_{i+1}  \\[3pt]
   	\mathbf{\tilde q}_{i+1}  \\
		\end{array}}
	\right] = {\left[ {
		\begin{array}{*{20}{c}}
   	\mathbf{\tilde J}_{\theta}(\mathbf{\tilde q}_{i},\mathbf{\tilde \theta }_{i}) \cdot \mathbf{\tilde K}_{\theta}^{-1}\cdot \mathbf{\tilde J}_{\theta}^T (\mathbf{\tilde q}_{i},\mathbf{\tilde \theta}_{i}) & \; \mathbf{\tilde J}_{q}(\mathbf{\tilde q}_{i},\mathbf{\tilde \theta}_{i})  \\ [3pt]
   	\mathbf{\tilde J}_{q}^T(\mathbf{\tilde q}_{i},\mathbf{\tilde \theta}_{i}) & \mathbf{0}  \\
		\end{array}} \right]^{ - 1}}
	\left[ {
		\begin{array}{*{20}{c}} 
   	\epsilon_{i}  \\[3pt]
   	0  \\
		\end{array}} 
	\right]
	\\ [14pt]

 	\mathbf{\tilde \theta}_{i+1} = \mathbf{\tilde K}_{\theta}^{-1}\; \cdot \mathbf{\tilde J}_{\theta }^T(\mathbf{\tilde q}_{i},\mathbf{\tilde \theta}_{i})\; \cdot \mathbf{F}_{i + 1}+ \mathbf{\tilde \theta}_{0} \\ [5pt]

	\epsilon_{i}=\mathbf{t} - \mathbf{g}(\mathbf{\tilde q}_{i},\mathbf{\tilde \theta}_{i}) + \mathbf{\tilde J}_{q}(\mathbf{\tilde q}_{i},\mathbf{\tilde \theta }_{i}) \cdot \mathbf{\tilde q}_{i} + \mathbf{\tilde J}_{\theta }(\mathbf{\tilde q}_{i},\mathbf{\tilde \theta}_{i}) \cdot (\mathbf{\tilde \theta}_{i}+\mathbf{\tilde \theta}_{0})  \\
	\end{array}
\end{equation}
where the starting point $({{\mathbf{\tilde{\theta }}}_{0}},{{\mathbf{\tilde{q}}}_{0}})$ is also computed iteratively, started from a nearest unloaded configuration where the joint coordinates are easily obtained from the inverse kinematic model. On the following iterations, to improve convergence, the system variables are slightly randomly disturbed. As follows from computational experiments, the proposed iterative algorithm possesses rather good convergence (3-5 iterations are usually enough).

\section{Stiffness matrix}

To compute the desired stiffness matrix, let us consider the neighborhood of the equilibrium configuration and assume that the external force and the end-effector location are incremented by some small values $\delta \mathbf{F}$, $\delta \mathbf{t}$. Besides, let us assume that a new configuration also satisfies the equilibrium conditions. Hence, it is necessary to consider simultaneously two equilibriums corresponding to the manipulator state variables $(\mathbf{F},\mathbf{q},\mathbf{\theta },\mathbf{t})$ and $(\mathbf{F}+\delta \mathbf{F},\mathbf{q}+\delta \mathbf{q},\mathbf{\theta }+\delta \mathbf{\theta },\mathbf{t}+\delta \mathbf{t})$. Relevant equations of statics may be written as

\begin{equation}\label{Eq:7}
  {\begin{array}{*{20}{l}}
  {{{\mathbf{\tilde J}}_\theta }^T {\mathbf{F}} = {{\mathbf{\tilde K}}_{\theta}} ({\mathbf{\tilde \theta}} - {{\mathbf{\tilde \theta}}_0});} &\;\;
    {{{\mathbf{\tilde J}}_q}^T {\mathbf{F}} = \mathbf{0};}  \\ [5pt]
    {{\left( {{{\mathbf{\tilde J}}_{\mathbf{\theta }}} + \delta {{\mathbf{\tilde J}}_{\mathbf{\theta }}}} \right)^T} \left( {{\mathbf{F}} + \delta {\mathbf{F}}} \right) = {{\mathbf{\tilde K}}_{\mathbf{\theta }}} \left( {{\mathbf{\tilde \theta }} - {{\mathbf{\tilde \theta}}_0} + \delta {\mathbf{\tilde \theta }}} \right);} & \;\; 
    {{\left( {{{\mathbf{\tilde J}}_{\mathbf{q}}} + \delta{{\mathbf{\tilde J}}_{\mathbf{q}}}} \right)^T} \left( {{\mathbf{F}} + \delta {\mathbf{F}}} \right) = \mathbf{0}} \\
\end{array}} 
 \end{equation}
where $\delta {{\mathbf{\tilde J}}_{q}}(\mathbf{\tilde q},\mathbf{\tilde \theta })$ and $\delta {{\mathbf{\tilde J}}_{\theta }}(\mathbf{\tilde q},\mathbf{\tilde \theta})$ are the differentials of the Jacobians due to changes in $(\mathbf{\tilde q},\,\,\mathbf{\tilde \theta })$. Besides, in the neighborhood of $(\tilde \mathbf{q},\mathbf{\tilde \theta })$, the kinematic equation (5) may be also  presented in the linearized form: 

\begin{equation}\label{Eq:8}
	\mathbf{\delta t}=\mathbf{\tilde J}_{\theta }(\mathbf{\tilde q},~\mathbf{\tilde \theta })\cdot \mathbf{\delta \theta }+\mathbf{\tilde J}_{q}(\mathbf{\tilde q},~\mathbf{\theta })\cdot \mathbf{\delta \tilde q},	
\end{equation}	

Hence, after neglecting the high-order small terms and expanding the differentials via the Hessians of the function $\Psi =\mathbf{\tilde g}{{(\mathbf{\tilde q},\mathbf{\tilde \theta })}^{T}}\mathbf{F}$

\begin{equation}\label{Eq:9}
\mathbf{\tilde H}_{qq}^{F}={\partial}^{2}\Psi /\partial \mathbf{\tilde q}^{2}; \;\;
\mathbf{\tilde H}_{\theta \theta }^{F}={\partial}^{2}\Psi /\partial \mathbf{\tilde \theta}^{2};\;\; 
\mathbf{\tilde H}_{q \theta}^{F}=({\mathbf{\tilde H}_{\theta q}^{F}})^{T}
={\partial}^{2}\Psi /\partial \mathbf{\tilde q}~\partial \mathbf{\tilde \theta },
\end{equation}
equations (7) may be rewritten as 

\begin{equation}\label{Eq:10}
\begin{array}{l}
 \mathbf{\tilde J}_{\theta}^T(\mathbf{\tilde q,\tilde \theta}) \cdot \delta \mathbf{F} + \mathbf{\tilde H}_{\theta q}^F(\mathbf{\tilde q,\tilde \theta}) \cdot \delta \mathbf{\tilde q} + \mathbf{\tilde H}_{\theta \theta}^F(\mathbf{\tilde q,\tilde \theta}) \cdot \delta \mathbf{\tilde \theta} = \mathbf{\tilde K}_{\theta} \cdot \delta \mathbf{\tilde \theta} \\ [5pt]
 \mathbf{\tilde J}_{q}^T(\mathbf{\tilde q,\tilde \theta}) \cdot \delta \mathbf{F} + \mathbf{\tilde H}_{qq}^F(\mathbf{\tilde q,\tilde \theta}) \cdot \delta \mathbf{\tilde q} + \mathbf{\tilde H}_{q\theta }^F(\mathbf{\tilde q,\tilde \theta}) \cdot \delta \mathbf{\tilde \theta} = \mathbf{0} \\ 
 \end{array}
 \end{equation}

Besides, here the variable $\delta \mathbf{\tilde \theta }$ can be eliminated analytically: $\delta \mathbf{\tilde \theta }=\mathbf{\tilde k}_{\theta }^{F}\cdot \mathbf{\tilde J}_{\theta }^{T}\cdot \delta \mathbf{F}+\mathbf{\tilde k}_{\theta }^{F}\cdot \mathbf{\tilde H}_{\theta q}^{F}\cdot \delta \mathbf{\tilde q}$, where $\,\mathbf{\tilde k}_{\theta }^{F}={{\left( {{\mathbf{\tilde K}}_{\theta }}-\mathbf{\tilde H}_{\theta \theta }^{\mathbf{F}} \right)}^{-1}}$. This leads to a system of matrix equations with unknowns  $\delta \mathbf{F}$ and $\delta \mathbf{\tilde q}$

\begin{equation}\label{Eq:11}
\left[ {\begin{array}{*{20}{c}}
   {\mathbf{\tilde J}_{\theta} \cdot \mathbf{\tilde k}_\theta ^F \cdot \mathbf{\tilde J}_{\theta}^T} & \;\; {\mathbf{\tilde J}_{\mathbf{q}} + \mathbf{\tilde J}_{\theta} \cdot \mathbf{\tilde k}_\theta ^F \cdot \mathbf{\tilde H}_{\theta q}^F}  \\ [3pt]
   {\mathbf{\tilde J}_{q}^{T} + \mathbf{\tilde H}_{q\theta }^{F} \cdot \mathbf{\tilde k}_{\theta} ^{F} \cdot \mathbf{\tilde J}_{\theta}^T} & \;\;
   \mathbf{\tilde H}_{qq}^{F} + \mathbf{\tilde H}_{q\theta}^{F} \cdot \mathbf{\tilde k}_{\mathbf{\theta}}^{F} \cdot \mathbf{\tilde H}_{\theta q}^{F}  \\
\end{array}} \right] \cdot \left[ {\begin{array}{*{20}{c}}
   \mathbf{\delta F}  \\ [3pt]
   \mathbf{\delta \tilde q}  \\
\end{array}} \right] = \left[ {\begin{array}{*{20}{c}}
   \mathbf{\delta t}  \\ [3pt]
   \mathbf{0}  \\
\end{array}} \right]
\end{equation}
from which the desired Cartesian stiffness matrix of the chain $\mathbf{K}_{c}$ may be obtained by direct inversion of the the left-hand side and extracting from it the upper-left sub-matrix of size $6\times6$:

\begin{equation}\label{Eq:12}
\left[ {\begin{array}{*{20}{c}}
   \mathbf{K}_{c} & \;\; {*}  \\
    {*}  &  \;\; {*}   \\
\end{array}} \right] = {\left[ {\begin{array}{*{20}{c}}
   \mathbf{\tilde J}_{\theta} \cdot \mathbf{\tilde k}_{\theta}^{F} \cdot \mathbf{\tilde J}_{\theta} ^{T} & \;\;
   \mathbf{\tilde J}_{q} + \mathbf{\tilde J}_{\theta} \cdot \mathbf{\tilde k}_{\theta}^{F} \cdot \mathbf{\tilde H}_{\theta q}^{F}  \\ [3pt]
   \mathbf{\tilde J}_{q}^{T} + \mathbf{\tilde H}_{q\theta }^{F} \cdot \mathbf{\tilde k}_{\theta} ^{F} \cdot \mathbf{\tilde J}_{\theta}^{T} & \;\;
   \mathbf{\tilde H}_{qq}^{F} + \mathbf{\tilde H}_{q\theta }^{F} \cdot \mathbf{\tilde k}_{\theta} ^{F} \cdot \mathbf{\tilde H}_{\theta q}^{F}  \\
\end{array}} \right]^{ - 1}}
\end{equation}

Finally, when the stiffness matrices  for all kinematic chains are computed, the stiffness of the entire multi-chain manipulator can be found by simple summation ${{\mathbf{K}}_{\Sigma}}=\sum\nolimits_{i=1}^{n}{{{\mathbf{K}}_{ci}}}$. It should be noted that, because of presence of the passive joints, the stiffness matrix of a separate serial kinematic chain is always singular, but aggregation of all the manipulator chains of a parallel manipulator produce a non-singular stiffness matrix.

\section{Kinetostatic control}

In robotics, the manipulator motions are usually generated using the inverse kinematic model that allows computing the input (reference) signals for actuators $\mathbf{\rho }$ corresponding to the desired end-effector location $\mathbf{t}$. However, for manipulators with preloaded passive joints, the kinematic control becomes non-applicable because of changes in the end-platform location due to the internal loading. Hence, in this case, the control must be based on the inverse kinetostatic model that takes into account both the manipulator geometry and elastic properties of its links and joints \cite{journal:Su}.

Using results from the previous sections, the desired inverse kinetostatic transformation can be performed iteratively, in the following way: 

\begin{description}

\item[\textbf{Step\#1.}]	For given target location of the end-platform $\mathbf{t}$, compute initial values of the actuated coordinates  ${{\mathbf{\rho }}_{0}}$  by applying the inverse kinematic transformation.

\item[\textbf{Step\#2.}]	For current values of the actuated coordinates  ${{\mathbf{\rho }}_{i}}$ and target location of the end-platform $\mathbf{t}$, find the equilibrium configuration for each kinematic chain and compute the corresponding total external loading $\mathbf{F}_{\Sigma }^{i}$ required to achieve the target location.

\item[\textbf{Step\#3.}]	If the computed external loading is less than the prescribed error, i.e.  $\left| \mathbf{F}_{\Sigma }^{i} \right|<{{\varepsilon }_{F}}$, stop the algorithm, otherwise continue the next step

\item[\textbf{Step\#4.}]	Repeat Step\#2 several times in the neighborhood of the current solution ${{\mathbf{\rho }}_{i}}$  and evaluate numerically the matrix $\mathbf{S}_{F\rho }^{i} = \partial \mathbf{F}_{\Sigma }^{i}/\partial \rho_{i}$ describing the sensitivity of $\mathbf{F}$ with respect to $\mathbf{\rho }$.

\item[\textbf{Step\#5.}]	Compute new value of the actuated coordinates $\mathbf{\rho}_{i+1}=\mathbf{\rho}_{i}-{\mathbf{S}_{F\rho}^{i}}^{-1}\cdot\mathbf{F}_{\Sigma}^{i}$ and repeat the algorithm starting from Step\#2.
\end{description}
 
As follows from simulation results, this algorithm demonstrates good convergence and can be used both for on-line and off-line trajectory planning. It was successfully applied to the case-study presented in the following Section.

\section{Application example}

Let us apply the proposed techniques to the stiffness analysis of the planar manipulator of the Orthoglide family (Figure 3). For illustration purposes, let us assume that the only source of the manipulator elasticity is concentrated in actuated drives, while the passive joints may be preloaded by (i) standard linear springs, or (ii) non-linear springs with mechanical stop-limit (see Figure 2 for details). 

\begin{figure}[t]
\center
\includegraphics[width=9.0cm]{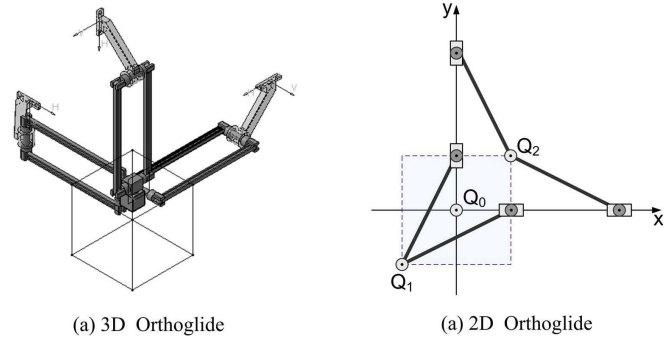}
\caption{Architecture of the Orthoglide manipulator and its planar version.}
\label{Figure:3}
\end{figure}

For this manipulator, the kinematic model includes a single parameter $L$ (the leg length) and the dexterous workspace was defined as the maximum square area that provides the velocity (and force) transmission factors in the range $[0.5,~2.0]$. Using the critical point technique developed for this type of manipulators \cite{journal:Chablat}, it was proved that the desired square vertices are located in the points ${Q_{1}(-p,~-p)}$ and ${Q_{2}(p,~p)}$, where $p=0.45~L$. Besides, the square centre ${Q_{0}(0,~0)}$ is isotropic with respect to the velocity and force transmission. The parameters of the actuating drives are also assumed identical and their linear stiffness is denoted as ${{K}_{\theta }}$. The auxiliary springs incorporated in the passive joints adjacent to the actuators are described by two parameters: the angular stiffness coefficient ${{K}_{\vartheta }}$ and the activation angle ${{\vartheta }_{0}}$ that defines the preloading activation point. During simulation, the manipulator end-point was displaced by value $\Delta$ in the direction $Q_{0}Q_{1}$ or $ Q_{0}Q_{2}$, and it was computed corresponding magnitude for external force $F$.

The stiffness analysis results are summarized in Figures 4, 5 and in Table 1. As follows from them, the original manipulator (without preloading in passive joints) demonstrates rather low stiffness in the neighborhood of the point $Q_2$, which is roughly 4 times lower than in the isotropic point $Q_0$. In contrast, the linear stiffness in the point $Q_1$ is twice higher than in the point $Q_0$. Besides, in the point $Q_2$, the external loading may provoke the buckling phenomenon that is caused by a local minimum of the force-deflection relation. In this case, the distance-to-singularity is essentially lower that it is estimated from the kinematical model and the manipulator may easily loose its structural stability.

\begin{figure}[t]
\center
\includegraphics[width=11.5cm]{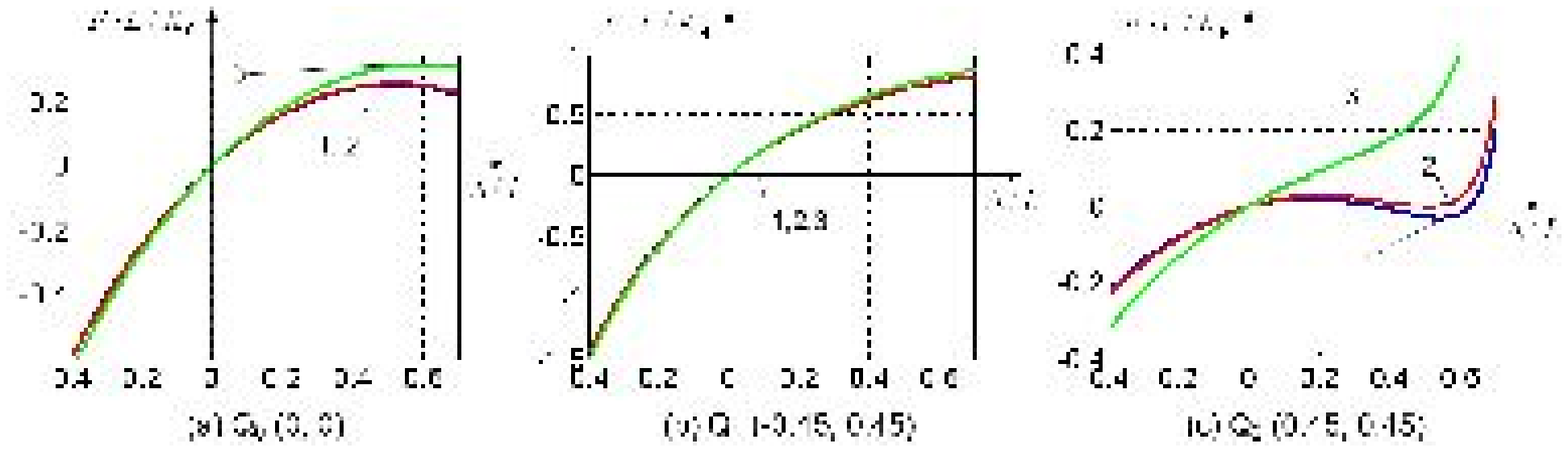}
\caption{Force-deflection relations $F=f(\Delta/L)$ in critical points:}
(1) $K_{\vartheta}=0$;~~~ (2) $K_{\vartheta} = 0.01~K_{\theta}~L^{2}$;~~~ (3) $K_{\vartheta} = 0.1~K_{\theta}~L^{2}$\\
(case of preloading with linear springs). 
\label{Figure:4}
\end{figure}

\begin{figure}[t]
\center
\includegraphics[width=11.5cm]{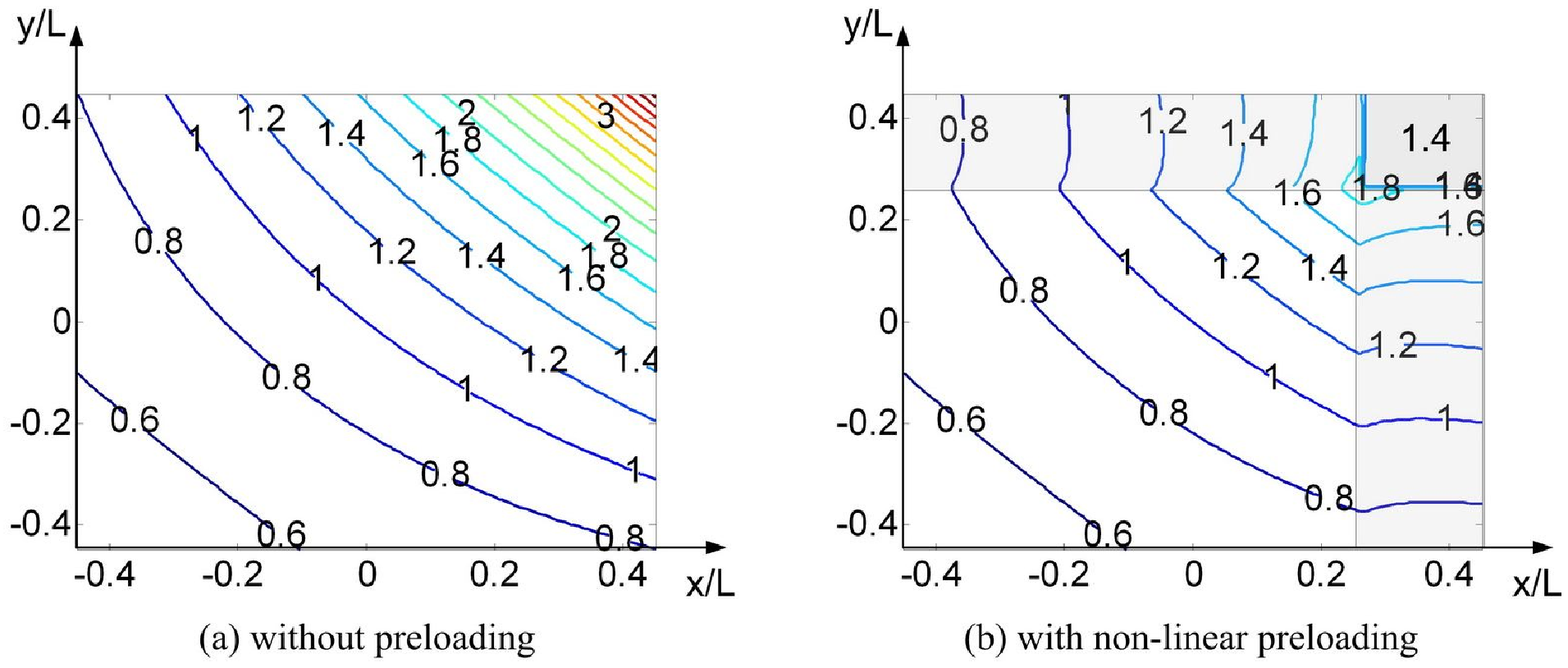}
\caption{Compliance maps for cases of: (a) manipulator without preloading;}
(b) manipulator with preloading non-linear springs with ${{K}_{\vartheta }}=0.5~{{K}_{\theta}}~{{L}^{2}}$ and ${{\vartheta }_{0}}=\pi/12$ (b).
\label{Figure:5}
\end{figure}

\begin{table}[t]
\caption{Manipulator stiffness for different linear preloading.}
\label{tri}
\begin{tabular}{p{3.8cm}p{1.8cm}p{1.8cm}p{1.8cm}p{1.8cm}}
\hline\noalign{\smallskip}
Stiffness in preloaded joints & $K_{\vartheta}=0$ & $0.01~K_{\theta}~L^{2}$ & $0.05~K_{\theta}~L^{2}$ & $0.1~K_{\theta}~L^{2}$ \\
\noalign{\smallskip}\svhline\noalign{\smallskip}
\multicolumn{5} {c} {Point $Q_{0}$ (isotropic point)} \\
\noalign{\smallskip}\svhline\noalign{\smallskip}
Actuating joint coordinates $\rho$ & $L$  & $L$ & $L$ & $L$ \\
Manipulator stiffness $\mathbf{K}_{c}$ & $K_{\theta}$ & $1.01~K_{\theta}$ & $1.05~K_{\theta}$& $1.10~K_{\theta}$\\
\noalign{\smallskip}\svhline\noalign{\smallskip}
\multicolumn{5} {c} {Point $Q_{1}$ (neighborhood of ``bar'' singularity)} \\
\noalign{\smallskip}\svhline\noalign{\smallskip}
Actuating joint coordinates $\rho$ & $0.437~L$  & $0.433~L$ & $0.419~L$ & $0.402~L$ \\
Manipulator stiffness $\mathbf{K}_{c}$ & $2.276~K_{\theta}$ & $2.286~K_{\theta}$ & $2.329~K_{\theta}$& $2.382~K_{\theta}$\\
\noalign{\smallskip}\svhline\noalign{\smallskip}
\multicolumn{5} {c} {Point $Q_{2}$ (neighborhood of ``flat'' singularity)} \\
\noalign{\smallskip}\svhline\noalign{\smallskip}
Actuating joint coordinates $\rho$ & $1.345~L$  & $1.356~L$ & $1.399~L$ & $1.453~L$ \\
Manipulator stiffness $\mathbf{K}_{c}$ & $0.24~K_{\theta}$ & $0.27~K_{\theta}$ & $0.39~K_{\theta}$ & $0.55~K_{\theta}$ \\
Critical force $\mathbf{F}_{cr}$ & $0.020~K_{\theta}~L$ & $0.027~K_{\theta}~L$ & --- & --- \\
\noalign{\smallskip}\hline\noalign{\smallskip}
\end{tabular}
\end{table}

To improve the manipulator stiffness and to avoid the buckling in the neighborhood of $Q_2$, the passive joints were first preloaded by linear springs with activation angle ${{\vartheta }_{0}}=0$. As follows from Figure 4, the preloading with parameter $K_{\vartheta}=0.1~K_{\theta}~L^{2}$ allows completely eliminate buckling and improves the stiffness by the factor of 2.3. On the other hand, the stiffness in the points $Q_0$ and $Q_1$ changes non-essentially, by 10\% and 5\% respectively. Hence, with respect to the stiffness, such preloading has positive impact.

The only negative consequence of such preloading is related to changes of the actuator control strategy. In fact, instead of standard kinematic control, it is necessary to apply the kinetostatic control algorithm presented in Section 5. It allows compensating the position errors caused by elastic deformations due to the internal preloading and to achieve the target end-point location with modified values of the actuated joint coordinates. As follows from Table 1, corresponding adjustments of the joint coordinates may reach $0.1~L$ and are not negligible for most of applications.

The most efficient solution that eliminates this problem is using of non-linear springs with mechanical stop-limits that are activated while approaching to $Q_2$. For instance, as follows from dedicated study, the preloading with the parameters  $K_{\vartheta}=0.5~{K_{\theta}~L}^{2}$, $\vartheta_{0}=\pi/12$ provides almost the same improvements in $Q_2$ as the linear spring while preserving usual control strategies if the preloading is not activated. The efficiency of this approach is illustrated by the compliance maps presented in Figure 5.

\section{Conclusions}

Recent advances in mechanical design of robotic manipulators lead to new parallel architectures that incorporates internal preloading in passive joints allowing to improve accuracy but leading to revision of existing stiffness analysis techniques. This paper presents new results in this area that allow simultaneously evaluate influence of internal and external loading and compute both the non-linear ``load-deflection'' relation and the stiffness matrices for any given location of the end-platform or actuating drives. Using this methodology, it is proposed the kinetostatic control algorithm that allows to improve accuracy of the classical kinematic control and to compensate position errors caused by elastic deformations in links/joints due to the external/internal loading. The efficiency of this technique is confirmed by an application example that deals with a parallel manipulator of the Orthoglide family where the internal preloading allows to eliminate the undesired buckling phenomena and to improve the stiffness in the neighborhood of its kinematic singularities. 

In future, these results will be generalized to other types of preloading that may be generated by external gravity-compensation mechanisms and also applied to micromanipulators with flexure joints.

\begin{acknowledgement}
The work presented in this paper was partially funded by the Region ``Pays de la Loire'' (project RoboComposite).
\end{acknowledgement}

\end{document}